\documentclass[11pt,a4paper]{article}
\usepackage[hyperref]{acl2019}
\usepackage{times}
\usepackage{latexsym}
\usepackage{url}
\usepackage{graphicx}
\usepackage{subcaption}
\usepackage{xcolor}
\usepackage{blindtext}

\aclfinalcopy 

\title{A Focus on Neural Machine Translation for African Languages}

\author{Laura Martinus \\
  Explore / Johannesburg, South Africa \\
  \texttt{laura@explore-ai.net} \\\And
 Jade Abbott \\
  Retro Rabbit / Johannesburg, South Africa \\
  \texttt{ja@retrorabbit.co.za} \\}

\date{}

\begin{document}
\maketitle
\begin{abstract}
African languages are numerous, complex and low-resourced. The datasets required for machine translation are difficult to discover, and existing research is hard to reproduce. Minimal attention has been given to machine translation for African languages so there is scant research regarding the problems that arise when using machine translation techniques. To begin addressing these problems, we trained models to translate English to five of the official South African languages (Afrikaans, isiZulu, Northern Sotho, Setswana, Xitsonga), making use of modern neural machine translation techniques. The results obtained show the promise of using neural machine translation techniques for African languages. By providing reproducible publicly-available data, code and results, this research aims to provide a starting point for other researchers in African machine translation to compare to and build upon.
\end{abstract}

\section{Introduction}


Africa has over 2000 languages across the continent \cite{enthnologue}. South Africa itself has 11 official languages. Unlike many major Western languages, the multitude of African languages are very low-resourced and the few resources that exist are often scattered and difficult to obtain.

Machine translation of African languages would not only enable the preservation of such languages, but also empower African citizens to contribute to and learn from global scientific, social and educational conversations, which are currently predominantly English-based \cite{alexander2010potential}. Tools, such as Google Translate \cite{google_translate}, support a subset of the official South African languages, namely English, Afrikaans, isiZulu, isiXhosa and Southern Sotho, but do not translate the remaining six official languages. 

Unfortunately, in addition to being low-resourced, progress in machine translation of African languages has suffered a number of problems. This paper discusses the problems and reviews existing machine translation research for African languages which demonstrate those problems. To try to solve the highlighted problems, we train models to perform machine translation of English to Afrikaans, isiZulu, Northern Sotho (N. Sotho), Setswana and Xitsonga, using state-of-the-art neural machine translation (NMT) architectures, namely, the Convolutional Sequence-to-Sequence (ConvS2S) and Transformer architectures. 

Section~\ref{problems} describes the problems facing machine translation for African languages, while the target languages are described in Section~\ref{languages}. Related work is presented in Section~\ref{relatedwork}, and the methodology for training machine translation models is discussed in Section~\ref{methodology}. Section~\ref{results} presents quantitative and qualitative results. 

\section{Problems}
\label{problems}

\begin{table*}[htbp!]
\centering
\small
\begin{tabular}{|l|l|l|l|l|l|}
\hline
Model                     & Afrikaans        & isiZulu   & N. Sotho & Setswana& Xitsonga      \\  \hline
\cite{google_translate} &41.18 & 7.54& & & \\
\cite{abbott2018towards}&  && &33.53 & \\ 
\cite{wilken2012developing}& && & 28.8&  \\ 
\cite{mckellar2014an}& & & & &37.31  \\ 
\cite{van2014exploring}&71.0  & 7.9 & 37.9 &&40.3   \\ \hline
\end{tabular}
\caption{BLEU scores for English-to-Target language translation for related work.}
\label{tab:related}
\end{table*}

The difficulties hindering the progress of machine translation of African languages are discussed below.

\textbf{Low availability} of resources for African languages hinders the ability for researchers to do machine translation. Institutes such as the South African Centre for Digital Language Resources (SADiLaR) are attempting to change that by providing an open platform for technologies and resources for South African languages \cite{bergh}. This, however, only addresses the 11 official languages of South Africa and not the greater problems within Africa.

\textbf{Discoverability}: The resources for African languages that do exist are hard to find. Often one needs to be associated with a specific academic institution in a specific country to gain access to the language data available for that country. This reduces the ability of countries and institutions to combine their knowledge and datasets to achieve better performance and innovations. Often the existing research itself is hard to discover since they are often published in smaller African conferences or journals, which are not electronically available nor indexed by research tools such as Google Scholar. 

\textbf{Reproducibility}: The data and code of existing research are rarely shared, which means researchers cannot reproduce the results properly. Examples of papers that do not publicly provide their data and code are described in Section~\ref{relatedwork}.

\textbf{Focus}: According to \citet{alexander2009evolving}, African society does not see hope for indigenous languages to be accepted as a more primary mode for communication. As a result, there are few efforts to fund and focus on translation of these languages, despite their potential impact.

\textbf{Lack of benchmarks}: Due to the low discoverability and the lack of research in the field, there are no publicly available benchmarks or leader boards to new compare machine translation techniques to. 

This paper aims to address some of the above problems as follows: We trained models to translate English to Afrikaans, isiZulu, N. Sotho, Setswana and Xitsonga, using modern NMT techniques. We have published the code, datasets and results for the above experiments on GitHub, and in doing so promote reproducibility, ensure discoverability and create a baseline leader board for the five languages, to begin to address the lack of benchmarks. 

\section{Languages}
\label{languages}

We provide a brief description of the Southern African languages addressed in this paper, since many readers may not be familiar with them. The isiZulu, N. Sotho, Setswana, and Xitsonga languages belong to the Southern Bantu group of African languages \cite{mesthrie2002language}. The Bantu languages are agglutinative and all exhibit a rich noun class system, subject-verb-object word order, and tone \cite{zerbian2007first}. N. Sotho and Setswana are closely related and are highly mutually-intelligible. Xitsonga is a language of the Vatsonga people, originating in Mozambique \cite{100yearsBill}. The language of isiZulu is the second most spoken language in Southern Africa, belongs to the Nguni language family, and is known for its morphological complexity \cite{keet2017grammar, BOSCH2017}. Afrikaans is an analytic West-Germanic language, that descended from Dutch settlers \cite{roberge2002afrikaans}.

\section{Related Work}
\label{relatedwork}

\begin{table*}[htbp!]
\centering
\small
\begin{tabular}{|l|r|r|r|r|r|}

\hline
Target Language               & Afrikaans & isiZulu & N. Sotho    & Setswana   & Xitsonga  \\ \hline
\# Total Sentences            &  53 172   & 26 728  &    30 777   &  123 868   &  193 587   \\ 
\# Training Sentences         &  37 219   &  18 709 &    21 543   &   86 706   &  135 510    \\ 
\# Dev Sentences              &  12 953   &  5 019  &     6 234   &   34 162   &   55 077  \\ 
\# Test Sentences             &   3 000   &   3 000 &     3 000   &    3 000   &    3 000  \\ 
\# Tokens                     & 714 103  & 374 860 &   673 200   & 2 401 206   & 2 332 713  \\ 

\# Tokens (English)           &  733 281   &  504 515 &  528 229 &  1 937 994  &  1 978 918  \\ \hline
\end{tabular}

\caption{Summary statistics for each dataset.}
\label{stats}
\end{table*}

\begin{table*}[th]
\centering
\small
\begin{tabular}{|p{4.0cm}|p{3.25cm}|p{3.2cm}|p{2.6cm}|}
\hline
Source & Target & Back Translation & Issue\\\hline
Note that the funds will be held against the Vote of the Provincial Treasury pending disbursement to the SMME Fund .& Lemali izohlala emnyangweni wezimali. & The funds will be kept at the department of funds . & Translation does not match the source sentence at all.\\\hline
Auctions & Ilungelo lomthengi lokwamukela ukuthi umphakeli unalo ilungelo lokuthengisa izimpahla & Consumer's right to accept that the supplier has the right to sell goods & Translation does not match the source sentence at all.\\\hline 
S E R V I C E S T A N D A R D S & AMAQOPHELO EMISEBENZI& & A space between each letter in the source sentence.\\
 \hline
\end{tabular}

\caption{Examples of issues pertaining to the isiZulu dataset.}
\label{tab:issues}
\end{table*}

This section details published research for machine translation for the South African languages. The existing research is technically incomparable to results published in this paper, because their datasets (in particular their test sets) are not published. Table \ref{tab:related} shows the BLEU scores provided by the existing work.

Google Translate \cite{google_translate}, as of February 2019, provides translations for English, Afrikaans, isiZulu, isiXhosa and Southern Sotho, six of the official South African languages. Google Translate was tested with the Afrikaans and isiZulu test sets used in this paper to determine its performance. However, due to the uncertainty regarding how Google Translate was trained, and which data it was trained on, there is a possibility that the system was trained on the test set used in this study as this test set was created from publicly available governmental data. For this reason, we determined this system is not comparable to this paper's models for isiZulu and Afrikaans.

\citet{abbott2018towards} trained Transformer models for English to Setswana on the parallel Autshumato dataset  \cite{groenewald2009introducing}. Data was not cleaned nor was any additional data used. This is the only study reviewed that released datasets and code. \citet{wilken2012developing} performed statistical phrase-based translation for English to Setswana translation. This research used linguistically-motivated pre- and post-processing of the corpus in order to improve the translations. The system was trained on the Autshumato dataset and also used an additional monolingual dataset. 

\citet{mckellar2014an} used statistical machine translation for English to Xitsonga translation. The models were trained on the Autshumato data, as well as a large monolingual corpus. A factored machine translation system was used, making use of a combination of lemmas and part of speech tags.

\citet{van2014exploring} used unsupervised word segmentation with phrase-based statistical machine translation models. These models translate from English to Afrikaans, N. Sotho, Xitsonga and isiZulu. The parallel corpora were created by crawling online sources and official government data and aligning these sentences using the HunAlign software package. Large monolingual datasets were also used. 

\citet{wolff2014experiments} performed word translation for English to isiZulu. The translation system was trained on a combination of Autshumato, Bible, and data obtained from the South African Constitution. All of the isiZulu text was syllabified prior to the training of the word translation system. 

It is evident that there is exceptionally little research available using machine translation techniques for Southern African languages. Only one of the mentioned studies provide code and datasets for their results. As a result, the BLEU scores obtained in this paper are technically incomparable to those obtained in past papers.

\section{Methodology}
\label{methodology}

The following section describes the methodology used to train the machine translation models for each language. Section~\ref{dataset} describes the datasets used for training and their preparation, while the algorithms used are described in Section~\ref{algorithms}.

\subsection{Data}
\label{dataset}
The publicly-available Autshumato parallel corpora are aligned corpora of South African governmental data which were created for use in machine translation systems \cite{groenewald2009introducing}. The datasets are available for download at the South African Centre for Digital Language Resources website.\footnote{Available online at: \url{https://repo.sadilar.org/handle/20.500.12185/404}} The datasets were created as part of the Autshumato project which aims to provide access to data to aid in the development of open-source translation systems in South Africa. 

The Autshumato project provides parallel corpora for English to Afrikaans, isiZulu, N. Sotho, Setswana, and Xitsonga. These parallel corpora were aligned on the sentence level through a combination of automatic and manual alignment techniques.

The official Autshumato datasets contain many duplicates, therefore to avoid data leakage between training, development and test sets, all duplicate sentences were removed.\footnote{Available online at: \url{https://github.com/LauraMartinus/ukuxhumana}} These clean datasets were then split into 70\% for training, 30\% for validation, and 3000 parallel sentences set aside for testing. Summary statistics for each dataset are shown in Table~\ref{stats}, highlighting how small each dataset is.

Even though the datasets were cleaned for duplicate sentences, further issues exist within the datasets which negatively affects models trained with this data. In particular, the isiZulu dataset is of low quality. Examples of issues found in the isiZulu dataset are explained in Table ~\ref{tab:issues}. The source and target sentences are provided from the dataset, the back translation from the target to the source sentence is given, and the issue pertaining to the translation is explained.

\begin{table*}[htbp!]
\centering
\small
\begin{tabular}{|l|l|l|l|l|l|}
\hline
Model                         & Afrikaans   & isiZulu   & N. Sotho & Setswana& Xitsonga  \\  \hline 

ConvS2S (Word)   & 16.17   & 0.28      & 7.41    & 24.18  & 36.96     \\ 
ConvS2S (Best BPE)& 25.04 (4k)        & 1.79 (4k) & 12.18 (4k)  &  26.36 (40k)   & 37.45 (20k)  \\ 
Transformer        & \textbf{35.26 (4k)}      & \textbf{3.33 (4k)}     & \textbf{24.16 (4k)}    & \textbf{28.07 (40k) }      & \textbf{49.74 (20k)}       \\ \hline
\end{tabular}
\caption{BLEU scores calculated for each model, for English-to-Target language translations on test sets.}
\label{tab:results}
\end{table*}

\subsection{Algorithms}
\label{algorithms}

We trained translation models for two established NMT architectures for each language, namely, ConvS2S and Transformer. As the purpose of this work is to provide a baseline benchmark, we have not performed significant hyperparameter optimization, and have left that as future work.

The Fairseq(-py) toolkit was used to model the ConvS2S model \cite{gehring2017convolutional}. Fairseq's named architecture ``fconv'' was used, with the default hyperparameters recommended by Fairseq documentation as follows: The learning rate was set to 0.25, a dropout of 0.2, and the maximum tokens for each mini-batch was set to 4000. The dataset was preprocessed using Fairseq's preprocess script to build the vocabularies and to binarize the dataset. To decode the test data, beam search was used, with a beam width of 5. For each language, a model was trained using traditional white-space tokenisation, as well as byte-pair encoding tokenisation (BPE). To appropriately select the number of tokens for BPE, for each target language, we performed an ablation study (described in Section~\ref{ablation}). 

The Tensor2Tensor implementation of Transformer was used \cite{tensor2tensor}. The models were trained on a Google TPU, using Tensor2Tensor's recommended parameters for training, namely, a batch size of 2048, an Adafactor optimizer with learning rate warm-up of 10K steps, and a max sequence length of 64. The model was trained for 125K steps. Each dataset was encoded using the Tensor2Tensor data generation algorithm which invertibly encodes a native string as a sequence of subtokens, using WordPiece, an algorithm similar to BPE \cite{kudo2018sentencepiece}. Beam search was used to decode the test data, with a beam width of 4.

\section{Results}
\label{results}

Section \ref{quantitative} describes the quantitative performance of the models by comparing BLEU scores, while a qualitative analysis is performed in Section \ref{qualitative} by analysing translated sentences as well as attention maps. Section \ref{ablation} provides the results for an ablation study done regarding the effects of BPE. 

\subsection{Quantitative Results}
\label{quantitative}

The BLEU scores for each target language for both the ConvS2S and the Transformer models are reported in Table \ref{tab:results}. For the ConvS2S model, we provide results for sentences tokenised by white spaces (Word), and when tokenised using the optimal number of BPE tokens (Best BPE), as determined in Section~\ref{ablation}. The Transformer model uses the same number of WordPiece tokens as the number of BPE tokens which was deemed optimal during the BPE ablation study done on the ConvS2S model.

In general, the Transformer model outperformed the ConvS2S model for all of the languages, sometimes achieving 10 BLEU points or more over the ConvS2S models. The results also show that the translations using BPE tokenisation outperformed translations using standard word-based tokenisation. The relative performance of Transformer to ConvS2S models agrees with what has been seen in existing NMT literature \cite{attention}. This is also the case when using BPE tokenisation as compared to standard word-based tokenisation techniques \cite{sennrich2015neural}. 

Overall, we notice that the performance of the NMT techniques on a specific target language is related to both the number of parallel sentences and the morphological typology of the language. In particular, isiZulu, N. Sotho, Setswana, and Xitsonga languages are all agglutinative languages, making them harder to translate, especially with very little data \cite{chahuneau2013translating}. Afrikaans is not agglutinative, thus despite having less than half the number of parallel sentences as Xitsonga and Setswana, the Transformer model still achieves reasonable performance. Xitsonga and Setswana are both agglutinative, but have significantly more data, so their models achieve much higher performance than N. Sotho or isiZulu. 

The translation models for isiZulu achieved the worst performance when compared to the others, with the maximum BLEU score of 3.33. We attribute the bad performance to the morphological complexity of the language (as discussed in Section~\ref{languages}), the very small size of the dataset as well as the poor quality of the data (as discussed in Section~\ref{dataset}). 

\subsection{Qualitative Results}
\label{qualitative}

We examine randomly sampled sentences from the test set for each language and translate them using the trained models. In order for readers to understand the accuracy of the translations, we provide back-translations of the generated translation to English. These back-translations were performed by a speaker of the specific target language. More examples of the translations are provided in the Appendix. Additionally, attention visualizations are provided for particular translations. The attention visualizations showed how the Transformer multi-head attention captured certain syntactic rules of the target languages. 
 
\subsubsection{Afrikaans}

\begin{figure*}[p]
\centering
\begin{subfigure}{0.5\textwidth}
  \centering\captionsetup{width=.8\linewidth}
  \includegraphics[width=.92\linewidth]{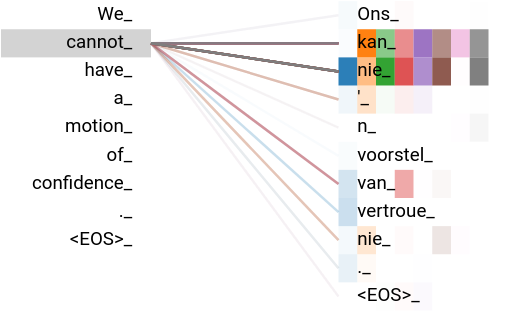}
  \caption{Visualization of multi-head attention for Layer 1 for the word ``cannot''. The coloured bars are individual attention heads. The word ``cannot'' is translated to ``kan nie ... nie'' where the second negative ``nie'' occurs at the end of the sentence.}
  \label{fig:sub1}
\end{subfigure}%
\begin{subfigure}{0.5\textwidth}
  \centering\captionsetup{width=.8\linewidth}
  \includegraphics[width=.9\linewidth]{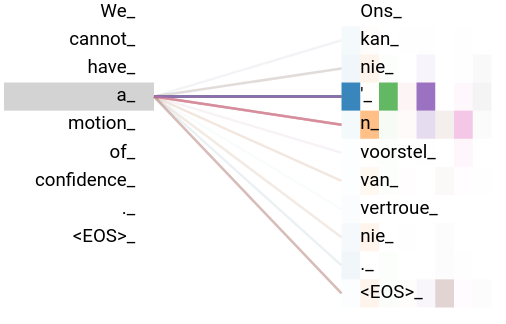}
  \caption{Visualization of multi-head attention for Layer 2 for the word ``a''. The coloured bars are individual attention heads. The word ``a'' is translated to ``'n'', as is successfully captured by the attention mechanism.}
  \label{fig:sub2}
\end{subfigure}
\caption{Visualizations of multi-head attention for an English sentence translated to Afrikaans using the Transformer model.}
\label{fig:afrikaansattention}
\end{figure*}

In Table~\ref{sent-result-table-afr}, ConvS2S did not perform the translation successfully. Despite the content being related to the topic of the original sentence, the semantics did not carry. On the other hand, Transformer achieved an accurate translation. Interestingly, the target sentence used an abbreviation, however, both translations did not. This is an example of how lazy target translations in the original dataset would negatively affect the BLEU score, and implore further improvement to the datasets. We plot an attention map to demonstrate the success of Transformer to learn the English-to-Afrikaans sentence structure in Figure~\ref{fig:afrikaansattention}.

\subsubsection{isiZulu}

Despite the bad performance of the English-to-isiZulu models, we wanted to understand how they were performing. The translated sentences, given in Table~\ref{sent-result-table-zulu}, do not make sense, but all of the words are valid isiZulu words. Interestingly, the ConvS2S translation uses English words in the translation, perhaps due to English data occurring in the isiZulu dataset. The ConvS2S however correctly prefixed the English phrase with the correct prefix ``i-". The Transformer translation includes invalid acronyms and mentions ``disease" which is not in the source sentence.  

\subsubsection{Northern Sotho}

If we examine Table~\ref{sent-result-table-sotho}, the ConvS2S model struggled to translate the sentence and had many repeating phrases. Given that the sentence provided is a difficult one to translate, this is not surprising. The Transformer model translated the sentence well, except included the word ``boithabišo'', which in this context can be translated to ``fun'' - a concept that was not present in the original sentence. 

\subsubsection{Setswana}

\begin{figure}[t]
    \centering
    \includegraphics[width=7cm]{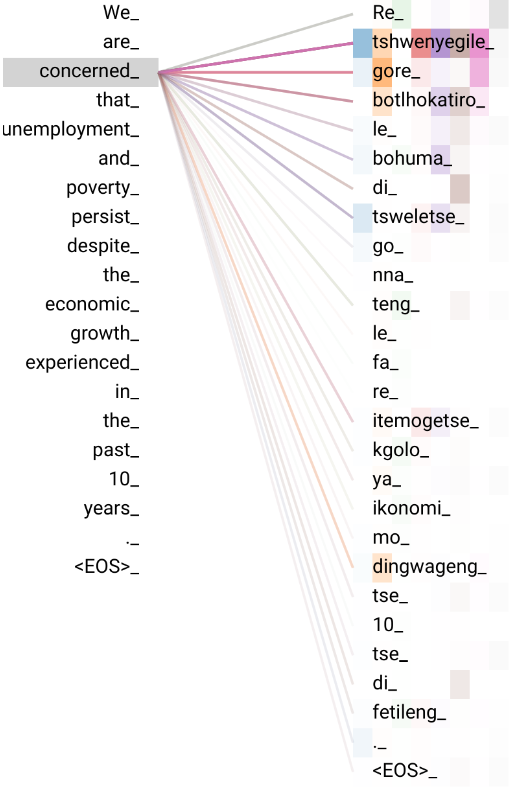}
    \caption{Visualization of multi-head attention for Layer 5 for the word ``concerned''. ``Concerned'' translates to ``tshwenyegile'' while ``gore'' is a connecting word like ``that''. }
    \label{fig:tswana}
\end{figure}

Table~\ref{sent-result-table-setswana} shows that the ConvS2S model translated the sentence very successfully. The word ``khumo'' directly means ``wealth'' or ``riches''. A better synonym would be ``letseno'', meaning income or ``letlotlo'' which means monetary assets. The Transformer model only had a single misused word (translated ``shortage'' into ``necessity''), but otherwise translated successfully. The attention map visualization in Figure~\ref{fig:tswana} suggests that the attention mechanism has learnt that the sentence structure of Setswana is the same as English.

\subsubsection{Xitsonga}

An examination of Table~\ref{sent-result-table-xitsonga} shows that both models perform well translating the given sentence. However, the ConvS2S model had a slight semantic failure where the cause of the economic growth was attributed to unemployment, rather than vice versa. 

\begin{table*}[p]

\caption{\textbf{English to Afrikaans Translations}: For the source sentence we show the reference translation, and the translations by the various models.  We also show the translation of the results back to English, performed by an Afrikaans
speaker.}
  \label{sent-result-table-afr}
  \centering
  \small
  \begin{tabular}{|p{3cm}|p{12cm}|}
  
    \hline
    
    Source  & \textcolor{blue}{Identity documents are issued to South African citizens or permanent residence permit holders who are 16 years or older.}\\
    Target & ID's word uitgereik aan Suid-Afrikaanse burgers en persone wat 'n permanente verblyfpermit het en 16 jaar oud of ouer is.\\\hline
    ConvS2S & Identiteitsdokumente word uitgereik aan Suid-Afrikaanse burgers of permanente verblyfpermit wat 16 jaar of ouer is.\\
    Back Translation & \textcolor{blue}{Identity documents are issued to South-African residents or permanent residence permits that are 16 years or older.}\\
    & \\
    Transformer & Identiteitsdokumente word aan Suid-Afrikaanse burgers of permanente verblyfhouers wat 16 jaar of ouer is, uitgereik.\\
    Back Translation &  \textcolor{blue}{Identity documents are issued to South-African residents or permanent residence holders that are 16 years or older.}\\\hline
    
  \end{tabular}
  \label{tab:sentences}
\end{table*}

\begin{table*}[p]

\caption{\textbf{English to isiZulu Translations}: For the source sentence, we show the reference translation, and the translations by the various models.  We also show the translation of the results back to English, performed by a isiZulu
speaker.}
  \label{sent-result-table-zulu}
  \centering
  \small
  \begin{tabular}{|p{3cm}|p{12cm}|}
  
    \hline
    
    Source  & \textcolor{purple}{Note that the funds will be held against the Vote of the Provincial Treasury pending disbursement to the SMME Fund .}\\
    Target & Lemali izohlala emnyangweni wezimali .\\\hline
    ConvS2S & Qaphela ukuthi izimali izokhokhela i-Vote of the Provincial Treasury ngokuthengiswa kwabe-SMME .\\
    Back Translation & \textcolor{purple}{Be aware the monies will pay the Vote of the Provincial Treasury with the paying by the SMME.}\\
    & \\
    Transformer & Qaphela ukuthi izimali zizobanjwa kweVME esifundazweni saseTreasury zezifo ezithunyelwa ku-MSE .\\
    Back Translation &  \textcolor{purple}{Be aware that the money will be held by VME with facilities of Treasury with diseases sent to MSE .}\\\hline
    
  \end{tabular}
  \label{tab:sentences}
\end{table*}

\begin{table*}[p]

\caption{\textbf{English to Northern Sotho Translations}: For the source sentence we show the reference translation, and the translations by the various models.  We also show the translation of the results back to English, performed by a Northern Sotho
speaker.}
  \label{sent-result-table-sotho}
  \centering
  \small
  \begin{tabular}{|p{3cm}|p{12cm}|}
  
    \hline
    
    Source  & \textcolor{blue}{No fishing vessel will be registered without a fishing right and a permit to engage in fishing activities .}\\
    Target & Ga go sekepe sa go rea dihlapi seo se tla go retšisetarwa/ngwadišwa ka ntle ga go ba le tokelo ya go rea dihlapi le tumelelo ya go kgatha tema mererong wa go rea dihlapi .\\\hline
    ConvS2S & Ga go phemiti ya go rea dihlapi e tla ngwadišwa ka ntle le phemiti ya go rea dihlapi le phemiti ya go rea dihlapi .\\
    Back Translation & \textcolor{blue}{There is no permit for fishing that can be registered for without a permit for fishing and a permit for fishing.}\\
    & \\
    Transformer & Ga go sekepe sa go rea dihlapi seo se tla go ngwadišwa tokelo ya go rea dihlapi le tumelelo ya go rea dihlapi ka mabaka a boithabišo. \\
    Back Translation & \textcolor{blue}{There is no fishing boat that will give you the right to fish and permission to fish for fun.}\\\hline
    
  \end{tabular}
  \label{tab:sentences}
\end{table*}

\begin{table*}[p]

\caption{\textbf{English to Setswana Translations}: For the source sentence we show the reference translation, and the translations by the various models.  We also show the translation of the results back to English, performed by a Setswana
speaker.}
  \label{sent-result-table-setswana}
  \centering
  \small
  \begin{tabular}{|p{3cm}|p{12cm}|}
  
    \hline

    Source  & \textcolor{purple}{If there is a shortage of bees in the sunflower lands during the flowering period it may lead to a possible loss of production by 25\% .}\\
    Target & Fa palo ya dinotshe e le kwa tlase mo masimong a disonobolomo ka nako ya go thuntsha dithunya go ka tlisa phokotso ya kumo e e ka nnang 25\% .\\\hline
    ConvS2S & Fa go na le tlhaelo ya dinotshe mo masimong a sonobolomo ka nako ya go thunya dithunya go ka tlisetsa tatlhegelo ya kumo ka 25\% \\
    Back Translation &  \textcolor{purple}{If there is a shortage of bees in the sunflower fields during the blooming season this can bring about a loss of income by 25\%}\\
    & \\
    Transformer & Fa go na le botlhokwa ba dinotshe mo masimong a disonobolomo fa nako ya go supa dithunya e ka simolola go thunya go ka fokotsa tatlhego ya kumo ka 25\% .\\ 
    Back Translation & \textcolor{purple}{If there is a need/importance/necessity/requirement/relevance for bees in the sunflower farms/fields when the blossoming season begins this can reduce loss of income by 25\%}\\\hline

  \end{tabular}
  \label{tab:sentences}
\end{table*}

\begin{table*}[p]
\caption{\textbf{English to Xitsonga Translations}: For the source sentence we show the reference translation, and the translations by the various models.  We also show the translation of the results back to English, performed by a Xitsonga
speaker.}
  \label{sent-result-table-xitsonga}
  \centering
  \small
  \begin{tabular}{|p{3cm}|p{12cm}|}
    \hline
    
    Source & \textcolor{blue}{we are concerned that unemployment and poverty persist despite the economic growth experienced in the past 10 years .}\\
    Target & hi na swivilelo leswaku mpfumaleko wa mitirho na vusweti swi ya emahlweni hambileswi ku nga va na ku kula ka ikhonomi eka malembe ya 10 lawa ya hundzeke .\\\hline
    ConvS2S & hi vilela leswaku ku pfumaleka ka mitirho na vusweti swi papalata ku kula ka ikhonomi eka malembe ya 10 lama nga hundza .\\
    Back Translation & \textcolor{blue}{We are concerned that the lack of jobs and poverty has prevented economic growth in the past 10 years.}\\
    & \\
    Transformer & hi na swivilelo leswaku mpfumaleko wa mitirho na vusweti swi ya emahlweni hambileswi ku nga va na ku kula ka ikhonomi eka malembe ya 10 lawa ya hundzeke .\\ 
    Back Translation & \textcolor{blue}{We have concerns that there is still lack of jobs and poverty even though there has been economic growth in the past 10 years.}\\\hline
   
  \end{tabular}
  \label{tab:sentences}
\end{table*}

\subsection{Ablation Study over the Number of Tokens for Byte-pair Encoding}
\label{ablation}

BPE \cite{sennrich2015neural} and its variants, such as SentencePiece \cite{kudo2018sentencepiece}, aid translation of rare words in NMT systems. However, the choice of the number of tokens to generate for any particular language is not made obvious by literature. Popular choices for the number of tokens are between 30,000 and 40,000: \citet{attention} use 37,000 for WMT 2014 English-to-German translation task and 32,000 tokens for the WMT 2014 English-to-French translation task. \citet{johnson2017google} used 32,000 SentencePiece tokens across all source and target data. Unfortunately, no motivation for the choice for the number of tokens used when creating sub-words has been provided.

\begin{figure}[htbp!]
    \centering
    \includegraphics[width=7.5cm]{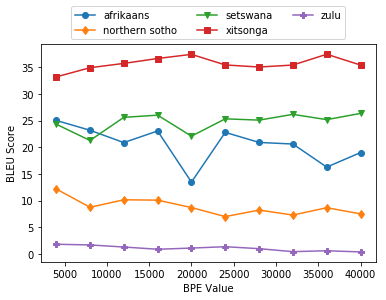}
    \caption{The BLEU scores for the ConvS2S of each target language w.r.t the number of BPE tokens.}
    \label{fig:convBPE}
\end{figure}

Initial experimentation suggested that the choice of the number of tokens used when running BPE tokenisation, affected the model's final performance significantly. In order to obtain the best results for the given datasets and models, we performed an ablation study, using subword-nmt \cite{sennrich2015neural}, over the number of tokens required by BPE, for each language, on the ConvS2S model. The results of the ablation study are shown in Figure~\ref{fig:convBPE}.

As can be seen in Figure~\ref{fig:convBPE}, the models for languages with the smallest datasets (namely isiZulu and N. Sotho) achieve higher BLEU scores when the number of BPE tokens is smaller, and decrease as the number of BPE tokens increases. In contrast, the performance of the models for languages with larger datasets (namely Setswana, Xitsonga, and Afrikaans) improves as the number of BPE tokens increases. There is a decrease in performance at 20 000 BPE tokens for Setswana and Afrikaans, which the authors cannot yet explain and require further investigation. The optimal number of BPE tokens were used for each language, as indicated in Table~\ref{tab:results}.  

\section{Future Work}
Future work involves improving the current datasets, specifically the isiZulu dataset, and thus improving the performance of the current machine translation models.

As this paper only provides translation models for English to five of the South African languages and Google Translate provides translation for an additional two languages, further work needs to be done to provide translation for all 11 official languages. This would require performing data collection and incorporating unsupervised  \cite{lample2017unsupervised, lample2019cross}, meta-learning \cite{gu2018meta}, or zero-shot techniques \cite{johnson2017google} .

\section{Conclusion}

African languages are numerous and low-resourced. Existing datasets and research for machine translation are difficult to discover, and the research hard to reproduce. Additionally, very little attention has been given to the African languages so no benchmarks or leader boards exist, and few attempts at using popular NMT techniques exist for translating African languages. 

This paper reviewed existing research in machine translation for South African languages and highlighted their problems of discoverability and reproducibility. In order to begin addressing these problems, we trained models to translate English to five South African languages, using modern NMT techniques, namely ConvS2S and Transformer. The results were promising for the languages that have more higher quality data (Xitsonga, Setswana, Afrikaans), while there is still extensive work to be done for isiZulu and N. Sotho which have exceptionally little data and the data is of worse quality. Additionally, an ablation study over the number of BPE tokens was performed for each language. Given that all data and code for the experiments are published on GitHub, these benchmarks provide a starting point for other researchers to find, compare and build upon.

The source code and the data used are available at \url{https://github.com/LauraMartinus/ukuxhumana}.

\section*{Acknowledgements}
The authors would like to thank Reinhard Cromhout, Guy Bosa, Mbongiseni Ncube, Seale Rapolai, and Vongani Maluleke for assisting us with the back-translations, and Jason Webster for Google Translate API assistance. Research supported with Cloud TPUs from Google's TensorFlow Research Cloud (TFRC).

\bibliography{acl2019}
\bibliographystyle{acl_natbib}
\newpage
\appendix
\onecolumn
\section{Appendix}
\label{sec:appendix}

Additional translation results from ConvS2S and Transformer are given in Table~\ref{app:sent-result-table} along with their back-translations for Afrikaans, N. Sotho, Setswana, and Xitsonga. We include these additional sentences as we feel that the single sentence provided per language in Section~\ref{qualitative}, is not enough demonstrate the capabilities of the models. Given the scarcity of research in this field, researchers might find the additional sentences insightful into understanding the real-world capabilities and potential, even if BLEU scores are low. 
\begin{table}[htbp!]

\caption{For each source sentence we show the reference translation, and the translations by the various models.  We also show the translation of the results back to English, performed by a home-language speaker.}
  \label{app:sent-result-table}
  \centering
  \small
  \begin{tabular}{|p{3cm}|p{12cm}|}
    \hline

    Source \textbf{Afrikaans} & \textcolor{blue}{If you want to work as a tourist guide in the Western Cape, you need to be registered with the provincial Tourist Guide Office.}\\
    Target  & As jy in die Wes-Kaap wil werk as 'n toergids, moet jy geregistreer wees by die provinsiale Toergidskantoor.\\\hline
    ConvS2S & As jy wil werk as 'n toergids in die Wes-Kaap, moet jy by die provinsiale Toeristeids geregistreer wees.\\ 
    Back Translation & \textcolor{blue}{If you want to work as a tour guide in the Western Cape, you have to be registered at the provincial $<$UNK$>$.}\\
    & \\
    Transformer & As jy wil werk as 'n toergids in die Wes-Kaap wil werk, moet jy geregistreer wees by die provinsiale Toergids kantoor.\\
    Back Translation & \textcolor{blue}{If you want to work as a tour guide in the Western Cape want to work, you have to be registered at the provincial Tour guide office.}\\\hline\hline
    
    Source \textbf{N. Sotho} & \textcolor{purple}{A veterinary import permit is needed to import animals or animal products into the Republic of South Africa .}\\
    Target & Tumelelo ya thekontle ya diphoofolo go di tlišwa ka nageng e a hlokagala go reka diphoofolo le ditšweletšwa tša diphoofolo ka ntle go tliša ka mo Repabliking ya Afrika Borwa .\\\hline
    ConvS2S & Tumelelo ya thekontle ya diphoofolo e a nyakega go reka diphoofolo tša diphoofolo goba ditšweletšwa tša diphoofolo ka ntle ga Afrika Borwa .\\
    Back Translation &\textcolor{purple}{Permission to import animals is needed to buy animals for animals products animals outside of South Africa.}\\
    & \\
    Transformer & Tumelelo ya thekontle ya diphoofolo go di tlišwa ka nageng e a hlokagala go reka diphoofolo le ditšweletšwa tša diphoofolo ka ntle go tliša ka mo Repabliking ya Afrika Borwa . \\
    Back Translation &\textcolor{purple}{Permission to import animals from outside the country is needed to buy animals and products for animals without bringing them into the Republic of South Africa.}\\\hline\hline
    
    Source \textbf{Setswana} & \textcolor{blue}{Councillors , together with officials , must ensure that there is enough money to deliver services and promote the planned development .}\\
    Target & Bakhanselara mmogo le baofisiri , ba tshwanetse go netefatsa gore go na le tšhelete e e lekaneng go rebola ditirelo le go tsweletsa tlhabololo e e rulagantsweng .\\\hline
    ConvS2S & Bakhanselara , mmogo le batlhankedi , ba tshwanetse go netefatsa gore go na le madi a a lekaneng a go rebola ditirelo le go rotloetsa tlhabololo e e rulagantsweng .\\
    Back Translation & \textcolor{blue}{Counselors, together with $<$UNK$>$, must ensure there is enough money to permit continuous services and to encourage development as planned.}\\
    & \\
    Transformer & Bakhanselara mmogo le batlhankedi , ba tshwanetse go netefatsa gore go na le madi a a lekaneng go rebola ditirelo le go tsweletsa tlhabololo e e rulagantsweng .\\ 
    Back Translation & \textcolor{blue}{Counselors together with $<$UNK$>$, are supposed to ensure that there's enough funds to permit services and to continue development as planned.}\\\hline\hline
    
    Source \textbf{Xitsonga}  & \textcolor{purple}{Improvement of performance in the public service also depends on the quality of leadership provided by the executive and senior management .}\\
    Target & Ku antswisa matirhelo eka mfumo swi tlhela swi ya hi nkoka wa vurhangeri lowu nyikiwaka hi vulawuri na vufambisi nkulu .\\\hline
    ConvS2S & Ku antswisiwa ka matirhelo eka mitirho ya mfumo swi tlhela swi ya hi nkoka wa vurhangeri lebyi nyikiwaka hi vufambisi na mafambiselo ya xiyimo xa le henhla .\\
    Back Translation & \textcolor{purple}{The improvement of the government’s work depends on the quality of the leadership which is provided by the management and the best system implemented.}\\
    & \\
    Transformer & Ku antswisiwa ka matirhelo eka vukorhokeri bya mfumo na swona swi ya hi nkoka wa vurhangeri lebyi nyikiwaka hi komitinkulu na mafambiselo ya le henhla . \\ 
    Back Translation & \textcolor{purple}{The improvement of service delivery by the government also depends on the quality of leadership which is provided by the high committee and the best implemented system.}\\\hline

  \end{tabular}
\label{app:sentences}
\end{table}

\end{document}